\documentclass[runningheads]{llncs}
\usepackage[T1]{fontenc}
\usepackage{boldline}
\usepackage{pifont}
\usepackage{bbm, dsfont}
\usepackage{amsmath}
\usepackage{amssymb}
\usepackage{amsfonts}
\usepackage{multirow}
\usepackage{booktabs}
\usepackage{hyperref}
\hypersetup{
    colorlinks=true,
}
\def\doi#1{\href{https://doi.org/\detokenize{#1}}{\url{https://doi.org/\detokenize{#1}}}}
\usepackage{graphicx}
\usepackage{color}
\usepackage{listings}

\def\ie{\textit{i.e.}}
\def\eg{\textit{e.g.}}

\begin{document}
\makeatletter\def\Hy@Warning#1{}\makeatother
\title{GL-Fusion: Global-Local Fusion Network for Multi-view Echocardiogram Video Segmentation}
\author{Ziyang Zheng\inst{1}$^{\dag\ddagger}$ \and
Jiewen Yang\inst{1}$^{\dag}$ \and
Xinpeng Ding\inst{1} \and 
Xiaowei Xu\inst{2}$^\star$ \and \\
Xiaomeng Li\inst{1}$^{\star}$}
\institute{The Hong Kong University of Science and Technology \and Guangdong Cardiovascular Institute, Guangdong Provincial People’s Hospital(Guangdong Academy of Medical Sciences), Southern Medical University ,Guangzhou, China}
%
\authorrunning{Z. Zheng$^\dag$; J. Yang$^\dag$; X. Ding; X. Xu$^\star$; X. Li$^\star$}
\maketitle              
\def\thefootnote{$\dag$}\footnotetext{Two authors contributed equally to this work.}\def\thefootnote{\arabic{footnote}}
\def\thefootnote{$\ddagger$}\footnotetext{Work completed during the internship at HKUST.}\def\thefootnote{\arabic{footnote}}
\def\thefootnote{$\star$}\footnotetext{Corresponding author: \email{eexmli@ust.hk, xiao.wei.xu@foxmail.com}}\def\thefootnote{\arabic{footnote}}
\begin{abstract}
Cardiac structure segmentation from echocardiogram videos plays a crucial role in diagnosing heart disease. 
The combination of multi-view echocardiogram data is essential to enhance the accuracy and robustness of automated methods.
However, due to the visual disparity of the data, deriving cross-view context information remains a challenging task, and unsophisticated fusion strategies can even lower performance. 
In this study, we propose a novel \textbf{G}obal-\textbf{L}ocal fusion (\textbf{GL-Fusion}) network to jointly utilize multi-view information globally and locally that improve the accuracy of echocardiogram analysis.
Specifically, a \textbf{M}ulti-view \textbf{G}lobal-based \textbf{F}usion \textbf{M}odule (MGFM) is proposed to extract global context information and to explore the cyclic relationship of different heartbeat cycles in an echocardiogram video.
Additionally, a \textbf{M}ulti-view \textbf{L}ocal-based \textbf{F}usion \textbf{M}odule (MLFM) is designed to extract correlations of cardiac structures from different views. 
Furthermore, we collect a multi-view echocardiogram video dataset (MvEVD) to evaluate our method. Our method achieves an 82.29\% average dice score, which demonstrates a 7.83\% improvement over the baseline method, and outperforms other existing state-of-the-art methods. To our knowledge, this is the first exploration of a multi-view method for echocardiogram video segmentation. Code available at: \href{https://github.com/xmed-lab/GL-Fusion}{https://github.com/xmed-lab/GL-Fusion}

\end{abstract}

\section{Introduction}
\label{sec:introduction}
Accurate segmentation of the cardiac structure from echocardiogram videos is integral to several analysis tasks~\cite{echo_seg1} and has a significant impact on clinical practice~\cite{echo_seg3}. 
For example, segmentation of the left ventricle (LV) enables quantifiable functional analysis of the heart, facilitating the detection and diagnosis of heart diseases~\cite{heart_disease1,heart_disease2,heart_disease3}.
Compared with the single view segmentation, multi-view information is crucial to diagnose heart disease,~\eg, the diagnosis of congenital heart disease requires the analysis of four views: parasternal long-axis view (PSLAX), parasternal short-axis view (PSSAX), subxiphoid long-axis view (SXLAX), and suprasternal long-axis view (SSLAX)~\cite{multiview_importance5}.
Consequently, to assist clinicians in diagnostic decision-making, there is a high demand for developing automated multi-view cardiac structure segmentation methods from echocardiogram videos in clinical practice.
\begin{figure}[t]
  \centerline{\includegraphics[width=\linewidth]{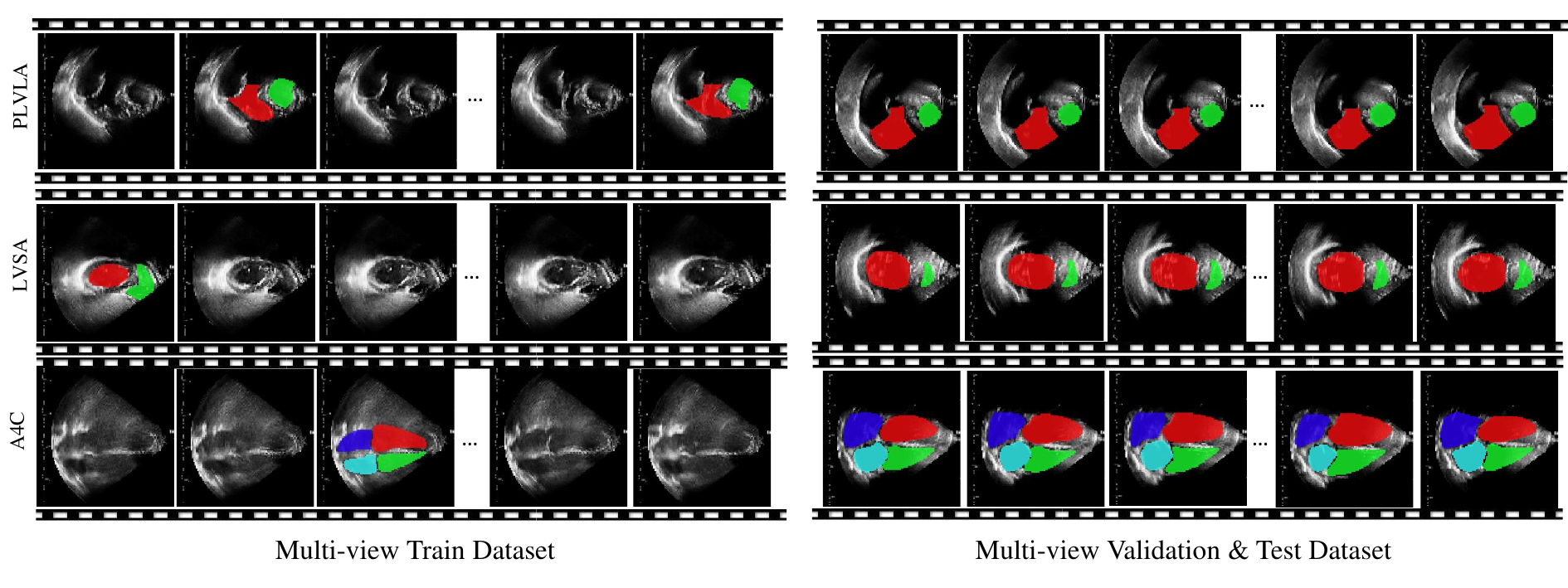}}
  \caption{Examples of multi-view echocardiogram dataset MvEVD, including PLVLA, LVSA, and A4C from top to bottom row. The colours red, green, blue, and cyan denote the LV, RV, LA, and RA cardiac structures. Our train set is sparsely annotated (5 frames per video), while the validation set and test set are fully annotated for each video frame.}
\vspace{-10pt}
\label{dataset}
\end{figure}
Existing echocardiogram segmentation approaches are primarily designed for single-view images or videos. 
For instance, Li et al.~\cite{echo_seg3} proposed a dynamic neural network capable of segmenting the LV from a long-axis fetal echocardiogram. In comparison, Leclerc et al.~\cite{echo_seg5} evaluated an encoder-decoder deep convolutional neural network that independently segments two and four-chamber images. 
However, these approaches have not addressed multi-view segmentation, where multi-view segmentation methods already exist in other medical domains, such as the CT-MRI~\cite{multiview13,multiview2,patel2016medical}, multi-view cardiac MRI~\cite{InfoTrans,rDLA,multiview9,TransFusion}, multi-view mammogram~\cite{multiview3}, and longitudinal multiple sclerosis~\cite{multiview8}.
Applying the proposed methods to multi-view echocardiogram segmentation presents several limitations: 
(1) Some methods are built for specific datasets and cannot adapt to our task. For instance, UMCT~\cite{multiview10} designated supervised training in one view by generating pseudo segmentation labels from other views, but has limitations in our task due to the significant gaps between views. In contrast, InfoTrans~\cite{InfoTrans} is designed for transmitting information between views instead of fusion them. While VCN~\cite{miccai_oral} employs contrastive learning to predict volume but may not be suitable for our task since defining positive and negative pairs is challenging due to the significant gap between views and labels.
(2) Methods such as JOIN~\cite{multiview3}, ROI-based fine-grained CNN~\cite{multiview4}, MIMTP~\cite{multiview8}, MV U-Net~\cite{multiview9}, MV-CNN~\cite{multiview11}, and Type-\uppercase\expandafter{\romannumeral1}, \uppercase\expandafter{\romannumeral2}, \uppercase\expandafter{\romannumeral3}~\cite{multiview13} concatenate the features or predicted probability maps of different views and then apply a fully-connected layer. However, these naive fusion strategies have shown limited performance and may even lead to worse results; see results in Table~\ref{comparsion}.
(3) Existing multi-view segmentation methods such as TransFusion~\cite{TransFusion} and rDLA~\cite{rDLA} mainly apply multi-view fusion with only global features. However, using global features for multi-view fusion may result in tangling the foreground/background pixels~\cite{local-attn} or leads to high levels of background noise in echocardiograms. 

To address this limitation, as shown in Fig.~\ref{dataset}, we first collect a multi-view echocardiogram video dataset, including three views: parasternal left ventricle long axis (PLVLA view), left ventricular short axis (LVSA) view, and apical 4 chamber (A4C) view. 
Different views of echocardiograms contain annotations for different chambers, such as, the PLVLA view contains the left ventricle (LV) and right ventricle (RV), the LVSA view contains the LV and RV, and the A4C view contains the LV, left atrium (LA), right atrium (RA), and RV. 
Furthermore, we propose a novel global-local fusion (GL-Fusion) network for multi-view echocardiogram video segmentation, where GL-Fusion includes a multi-view local-global fusion module designed to aggregate information from different views and improve the representation of each view.
The GL-Fusion comprises two components. First, a multi-view global fusion module (MGFM) interacts with the global semantics between different views and thus enhances the representation of each view. 
Second, since the global semantics may contain a significant amount of noisy information, a multi-view local fusion module (MLFM) is introduced to encourage the model to focus on foreground information.

In addition to capturing multi-view information, we propose a novel dense cycle loss designed to utilize unlabelled video data for improved representation learning. 
Our motivation is based on the idea that standard multi-view data is obtained from the same patient and under the same stable conditions, without abnormal behaviours such as suffocating or exercising, ensuring consistent cardiac cycles.
Previous work~\cite{weihang} proposed an unsupervised method called cycle loss, which trains the model with unlabelled frames based on the heartbeat cycle's characteristics. 
Nevertheless, the proposed cycle loss only focuses on a pair in two different cycles but ignores possibly similar images that may appear simultaneously in a systolic or diastolic period, resulting in features from similar frames being considered distant. 
To address this issue, our dense cycle loss examines all possible pairings throughout the heartbeat cycle. In summary, our contributions are as follows:
\begin{itemize}
    \item To the best of our knowledge, this is the first study to examine multi-view echocardiogram video segmentation.
    \item Our proposed GL-Fusion uses a multi-view local-global fusion module  to combine information from different views and improve the representation of each view.
    \item We further design a dense cycle loss that utilizes unlabelled data to enforce feature similarity based on temporal cyclicality.
    \item Extensive experiments demonstrate our method improved performance over existing methods, achieving an average dice score of 0.81. We plan to make our code publicly available upon paper acceptance.
\end{itemize}
\begin{figure}[!ht]
  \centerline{\includegraphics[width=\linewidth]{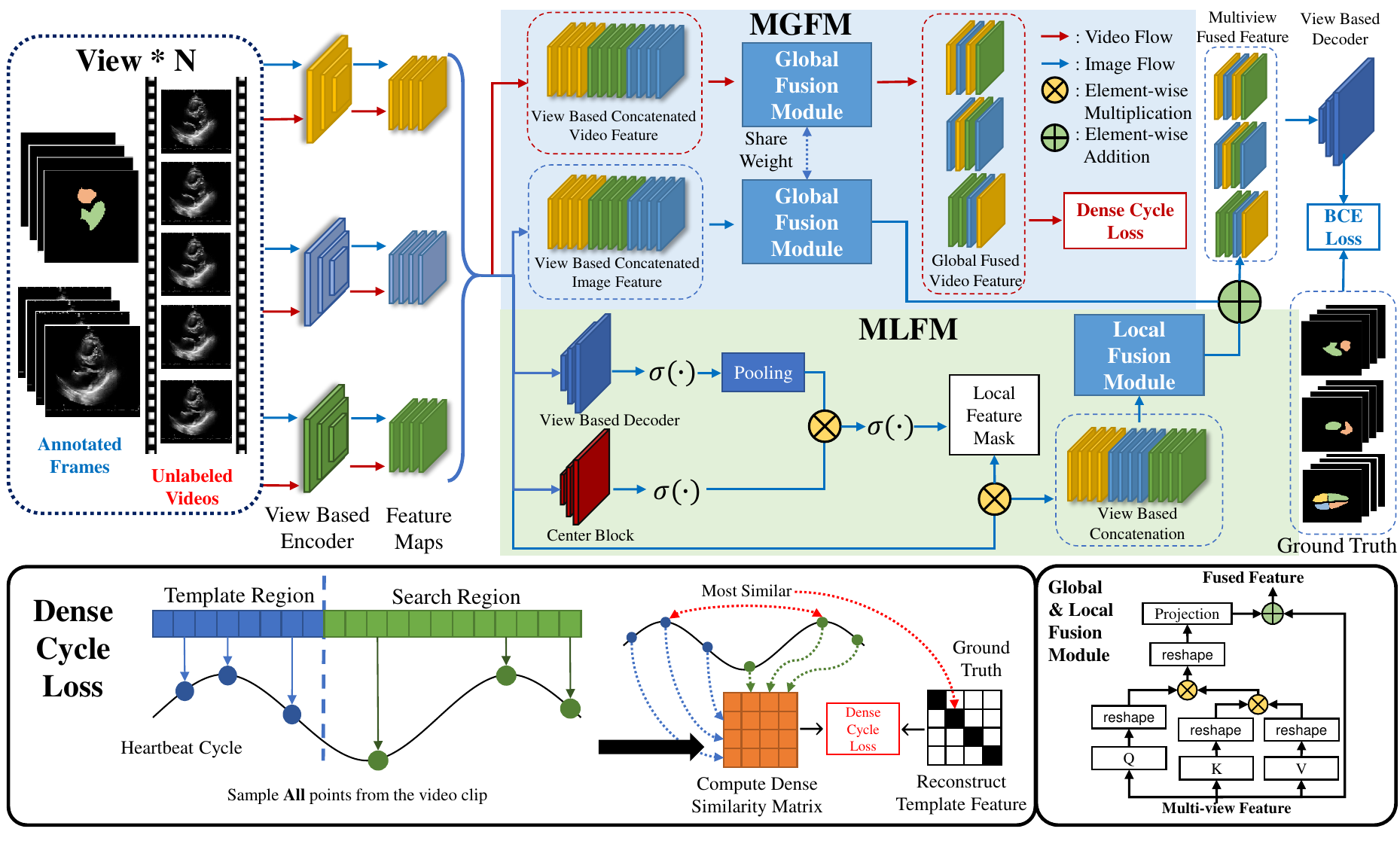}}
  \caption{\textbf{The overview framework of GL-fusion}. The Multi-view Global-based Fusion Module (MGFM) is proposed for global context information extraction and introduces dense cycle loss to devise the enforcement of the similarity of dense features between two heartbeat cycles from an echocardiogram video. The proposed Multi-view Local-based Fusion Module (MLFM) focuses on mining the correlation of local features of chambers in a different view.
  }
\label{framework}
\vspace{-5pt}
\end{figure}
\section{Methodology}
\label{method}
\subsection{The Overall Framework}
Fig.~\ref{framework} shows the overall pipeline of our proposed Multi-view Echocardiogram Global-Local Fusion Network (GL-Fusion), which consists of four main components: a view-based encoder, a multi-view global-local fusion module, a dense cycle loss module and a view-based decoder, where view-based indicate that parameters of the network of each independent view are non-shared.
In our experiment, we use DeeplabV3\cite{Deeplabv3} as our view-based encoder and decoder.
Formally, we denoted the sample echocardiogram videos as $\mathbf{V}=\{ \mathbf{X}^{i} \}_{i=1}^V$, where $\mathbf{X}^{i}\in\mathbb{R}^{C\times H\times W\times T}$ is the $i$-th view video and $V$ is the number of views, and, $C$, $H$, $W$ and $T$ indicate the channels, height, width, and length of input images.
Each video consists of $T$ frames,~\ie, $\mathbf{X}^{i} = \{ \mathbf{x}^i_{t} \}_{t=1}^T$, where $T$ remain the same for different view and $\mathbf{x}^i_{t}\in\mathbb{R}^{C\times H\times W}$ indicate $t$-th frame of $i$-th view video.
Since only sparse frames are provided segmentation annotation for training in a video, thus we denote the annotation frame pair as $\{ \mathbf{x}^i_{t_n},\mathbf{y}^i_{t_n} \}_{n=1}^N$, where $t_n$ is the index of the annotation and $N$ is the number of labelled frames that $N<<T$. 

During the training, We feed the videos $\mathbf{V}$ into the view-based encoder to extract the corresponding feature maps $\{ {\mathbf{F}}^i \}_{i=1}^V$ of each view, where  $\mathbf{F}^i\in\mathbb{R}^{D\times h\times w\times T}$, and, $D$, $h$ and $w$ indicate the channel number, height and width of feature maps.
Then the multi-view global-local fusion module aims to obtain the multi-view fused features $\{ \overline{\mathbf{F}}^i\}_{i=1}^V$, which extract global and local semantics information from other views to enhance the representation of each view (See Section~\ref{fusion}).
Following is the view-based decoder that generates the predicted segmentation result $\mathbf{y}^i$ from fused features, and maps the results to corresponding segmentation annotation,~\ie, $ \hat{\mathbf{y}}^i_{t_n}$ to the segmentation masks ${\mathbf{y}}_{t_n}^{i}$.
For the annotated frames, we use the segmentation loss to supervise them, formulated as follows:
\begin{equation}
   \mathcal{L}_{seg} = \sum_{i=1}^V \sum_{t_n=1}^{N}\mathcal{L}_{bce}(\hat{\mathbf{y}}_{t_n}^{i}, \mathbf{y}^i_{t_n}),
\end{equation}
where $\mathcal{L}_{bce}$ is the Binary Cross Entropy.
The sparse annotations are only a few frames in the whole video; thus can not obtain a robust model.
To leverage a large number of unlabelled frames, we design the dense cycle loss $\mathcal{L}_{cyc}$ to enforce temporal feature similarity of videos based on cyclicality; See Section~\ref{cycle_loss}.
The overall loss function of our model is as follows:
\vspace{-3pt}
\begin{equation}
    \mathcal{L} = \mathcal{L}_{seg} + \alpha \mathcal{L}_{cyc},
\vspace{-3pt}
\end{equation}
where $\alpha$ is the hyper-parameter to control the weight between two losses.
In the following, we will illustrate the multi-view global-local fusion module and the dense cycle loss in detail.
\vspace{-10pt}
\subsection{Multi-view Global-local Fusion Module}
\label{fusion}
In this section, we describe the multi-view global-local fusion module that aggregates the information from different views to enhance their feature representation.
To this end, we first concatenate extracted feature $\{ {\mathbf{F}}^i \}_{i=1}^V$ from different views in a view-wise manner to obtain $\mathbf{F} = \{ \mathbf{f}_t \}_{t=1}^T$, where $\mathbf{f}_t$ is the $t$-th feature vector in $\mathbf{F}$, and $\mathbf{f}_t \in \mathbb{R}^{D \times  V \times w \times h}$.
Then, we describe the multi-view global and local fusion with $\mathbf{F}_{global}$ and $\mathbf{F}_{local}$, respectively.

\vspace{2.5mm}
\noindent \textbf{Multi-view global fusion.} 
In order to enhance the representation of each view, we propose the global-based fusion module (MGFM)  to interact with the global semantics between different views.
To this end, we introduce a view-wise non-local block, which extracts the context information across views.
Similarly to the previous research~\cite{non-local,ding2021support} that applied attention to fuse the information, we here introduce the view-wise attention module to aggregate the cross-view information (see Figure~\ref{framework}). 
Then fused feature $\overline{\mathbf{F}}_{global}$ will be sent to both compute the dense cycle loss and cooperate with the local fused feature for segmentation prediction.

\vspace{2.5mm}
\noindent \textbf{Multi-view local fusion.} 
Since each view represents different morphological information of the heart and may contain the same cardiac structure as others, for example, the view PLVLA and LVSA both contain left ventricle(LA) and right ventricle(RV). 
Hence, extracting the local feature that represents the cardiac structure can contribute to feature fusion more efficiently. 
In this module, the extracted feature $\mathbf{F}_{local}$ will first pass to both the view-based decoder and a center block, where the decoder and center block has the same components with different output. 
The decoder provides the pseudo label $\{\hat{\mathbf{y}}^i\}^V_{i=1}$ of different cardiac structures. 
A center block is introduced to acquire the weight $\{w^i\}^V_{i=1}$ of $\{\hat{\mathbf{y}}^i\}^V_{i=1}$ and compute the local feature masks $\{\mathcal{M}^i\}^V_{i=1}$ as equation~\ref{eq:feature_mask},
\begin{equation}
    \mathcal{M}^i=\sigma(pooling(\sigma(\hat{\mathbf{y}}^i)) \times \sigma(w^i)),
    \label{eq:feature_mask}
\end{equation}
where weight $w$ has the greatest volume in the central area of the segmented regions and attenuation with distance, $\sigma$ denotes the sigmoid function and $\mathcal{M}\in \mathbb{R}^{1 \times H \times W\times T}$.
These masks highlight features with a stronger intensity that are closer to the object center, while discarding background information that is farther away from the center. 
This selection is based on the understanding that morphological information should remain consistent closer to the center.
In the final, similar to the process of MLFM, the view-wise local feature will be conducted view-wise concatenation operation and multiplied with local feature mask $\{\mathcal{M}^i\}^V_{i=1}$. 
Then sent to the view-wise attention module to acquire the local fused feature $\overline{\mathbf{F}}_{local}$.
\vspace{-6pt}
\subsection{Dense Cycle Loss}
\label{cycle_loss}
In echocardiogram videos, since only sparse annotation is available for the supervised training, involving the unlabelled data for our training and enhancing the performance is a challenge. 
The previous research~\cite{weihang} proposes an unsupervised method named cycle loss, which jointly trains the model with the unlabelled frames according to the characteristic of the heartbeat cycle. 
However, the proposed cycle loss considers only one clip in an iteration, which has the possibility to match frames that are morphologically identical but not in the same state, such as the search region being end-diastole while the template region is end-systole.

Thus, we propose the dense cycle loss, which considers all the possible matching across all template and search regions in each view independently.
For the multi-view fused feature $\overline{\mathbf{F}}_{global}$ of each video will be separated to template region $P^i$ and search region $Q^i$ with a ratio in 2:3 according to total frame length $T$.
Then we densely sample all feature intervals $\{p_1^i,...,p_n^i\}$ from $P^i$ and $\{q_1^i,...,q_m^i\}$ from $Q^i$, respectively, both sampling use the same chunk size $s$ and in our experiment, $n$ and $m$ is $\frac{2}{5}\times \frac{T}{s}$ and $\frac{3}{5}\times \frac{T}{s}$. 
Then we compute the similarity between candidate interval $p_k^i$ and target intervals $q_j^i$ of $Q^i$. 
\vspace{-3pt}
\begin{equation}
    \alpha^i_{j} = \sum{\mathcal{W}(\{p\}_k^i,\{q\}_j^i)}\times\{q\}_j^i,
\vspace{-3pt}
\end{equation}
where $\mathcal{W}(\cdot)$ is the computation of the similarity matrix. The similarity will be used as the weight to reconstruct the feature interval $\tilde{p}_k^i$.
Then we back to template region $P^i$ and compute the similarity between $\tilde{p}_k^i$ and all feature intervals$\{p_1^i,...,p_{n}^i\}$ in $P^i$.
Then we consider the index of $p_k^i$ as one-hot label $g$ of the most similar interval of $\tilde{p}_k^i$ and compute view-wise cycle loss $\mathcal{L}_{cyc}$ with label $g$ as shown in the following equation:
\begin{equation}
    \mathcal{L}_{cyc} = \sum_{i=1}^{V}\sum_{j\in P^i} \mathds{1}_{j=g}log(\alpha^i_{j})
\end{equation}
\section{Experiment}
\label{dataset_section}
\noindent \textbf{Datasets.} 
We collect a large multi-view echocardiogram video dataset named \textbf{MvEVD} from one medical institution, with a total of 254 sparsely annotated videos and 10 fully annotated videos with $800 \times 600$ resolution across three cardiac views (PLVLA, LVSA and A4C view). Each video includes 5 annotated frames. The average length of each video is larger than $100$ frames that are able to cover more than one cardiac cycle.

\noindent \textbf{Implementation Details.} We use the model DeeplabV3~\cite{Deeplabv3} as our view-based encoder and decoder, and select Adam optimizer for the model training with initial learning rate as $3e^{-4}$ and weight decay of $1e^{-5}$. When training, we use all sparsely annotated videos. All annotated frames are selected to supervise training while randomly selecting $40$ consecutive frames from videos for semi-supervised training. The training batch size of annotated images and unlabeled videos is 8 and 1, respectively.
In the final, we use CosineAnnealing as a scheduler and set the total training epoch to 100. The framework is built with Pytorch with 4 NVIDIA RTX3090 GPUs for training.
For the data augmentation in the training stage, we resize each frame in $144 \times 144$ size and then randomly crop them to $112 \times 112$. 

\noindent\textbf{Validation and testing.}
We use all fully annotated videos and split them into validation and testing with a ratio of 2:8. In this stage, we resize each frame in $144 \times 144$ size and conduct center cropping to them with the size of $112 \times 112$. Selecting the best model based on validation performance and report results in the testing set with Dice score. 
%
\begin{table}[!t]
\caption{The comparison with other methods. all results are reported in Dice Score.}
\renewcommand{\arraystretch}{1.2}
\setlength{\tabcolsep}{6pt}
\begin{tabular}{c|c|ccc|c}
\hlineB{3}
                                & Method        & PLVLA          & LVSA           & A4C            & Average Dice (\%)   \\ 
                                \hline
                                \hline
\multirow{3}{*}{Single-view} & DeeplabV3~\cite{Deeplabv3}     & 70.93          & 75.14          & 77.33          & 74.46          \\
                                & U-Net~\cite{u-net}         & 73.35          & 77.57          & 76.60          & 75.84          \\
                                & CSS~\cite{weihang}           & 79.09          & 79.70          & 77.71          & 78.83          \\ \hline
\multirow{4}{*}{Fusion-based}   & Early-fusion  & 79.78          & 77.07          & 77.58          & 78.14          \\
                                & Mid-fusion    & 77.89          & 76.75          & 72.44          & 75.69          \\
                                & Late-fusion   & 71.62          & 75.31          & 74.68          & 73.87          \\ & TransFusion~\cite{TransFusion}           & 78.79         & 80.23          & 59.31          & 72.78          \\
                                & \textbf{Ours} & \textbf{83.84} & \textbf{81.76} & \textbf{81.28} & \textbf{82.29} \\ \hlineB{3}
\end{tabular}
\label{comparsion}
\end{table}
\begin{table}[t]
    \begin{minipage}[t]{.479\linewidth}
        \caption{\textbf{Effectiveness of MGFM and MLFM.} This table shows the performance of the Global and Local fusion modules}
        \setlength{\tabcolsep}{0pt}
        \begin{center}
        \resizebox{0.99\textwidth}{!}{
            \begin{tabular}{c|c|c|c}
            \hlineB{3}
                      & MGFM & MLFM & Avg. Dice(\%)   \\ \hline\hline
            Base      & \ding{55}    & \ding{55}    & 74.46            \\
            Base+MGFM & \checkmark    & \ding{55}    & 80.20           \\
            Base+MLFM & \ding{55}    & \checkmark    & 78.41           \\
            Ours      & \checkmark    & \checkmark    & \textbf{82.29} \\ \hlineB{3}
            \end{tabular}
        }
        \end{center}
    \end{minipage}
    \hspace{5pt}
    \begin{minipage}[t]{.479\linewidth}
        \caption{\textbf{Effectiveness of Cyc. and Dense Cyc..} This table shows the effectiveness of vanilla cycle loss~\cite{weihang} (Noted by Cyc.) and our proposed dense cycle (Noted by Dense Cyc.).}
        \setlength{\tabcolsep}{0pt}
        \begin{center}
        \vspace{-3pt}
        \resizebox{0.99\textwidth}{!}{
            \begin{tabular}{c|c|c|c}
            \hlineB{3}
                        & Cyc.          & Dense Cyc.   & Avg. Dice(\%)       \\ \hline\hline
            Fusion-only      & \ding{55}     & \ding{55}    & 80.36          \\
            Fusion+Cyc.       & \checkmark    & \ding{55}    & 79.33          \\
            GL-Fusion          & \checkmark    & \checkmark   & \textbf{82.29} \\ \hlineB{3}
            \end{tabular}
        }
        \end{center}
    \end{minipage}
\end{table}
\subsection{Comparison with the State-of-the-Art Methods}
\begin{figure}[!t]
\centerline{\includegraphics[width=\linewidth]{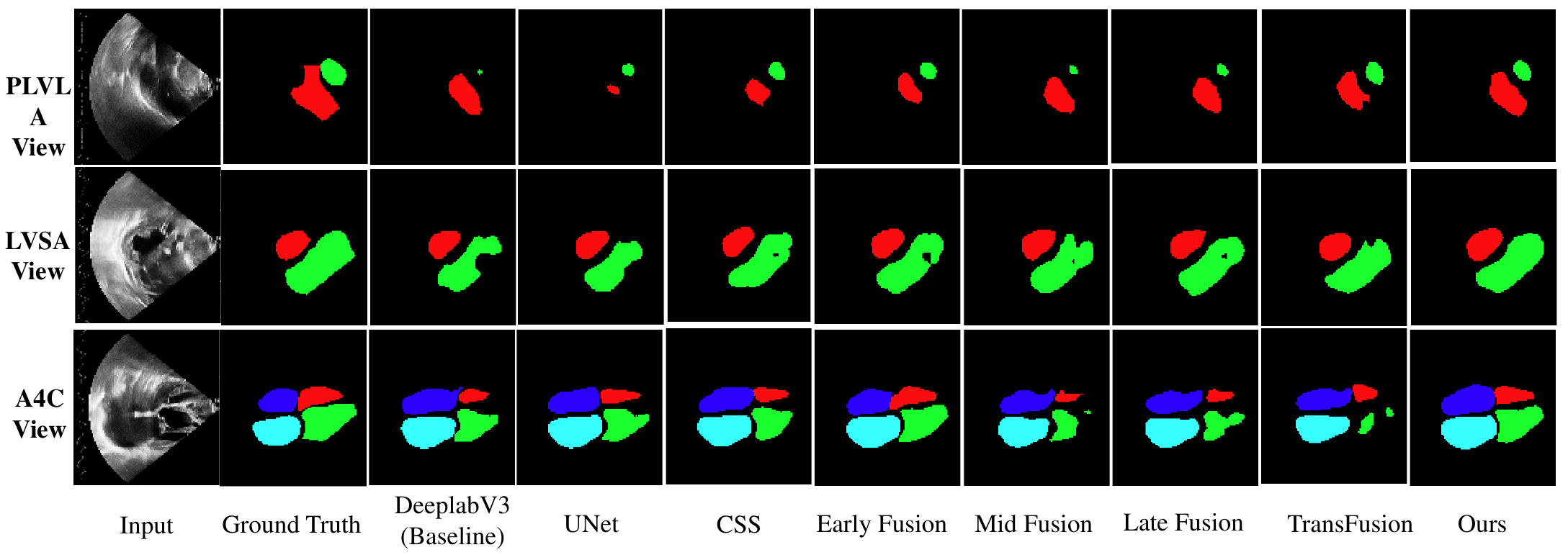}}
    \caption{Segmentation results from three views of echocardiogram videos, including PLVLA, LVSA, and A4C from top to bottom row. The red, green, blue, and cyan colours refer to LV, RV, LA, and RA cardiac structures, respectively.}
\label{visualize}
\end{figure}

To evaluate the performance of our method, we do the comparison with two types of methods: single-view methods and fusion-based methods in Table~\ref{comparsion}.
To be specific, single-view methods independently train segmentation networks for each view without using any strategy across views or simply conducting semi-supervised approaches~\cite{weihang}.
Fusion-based methods use feature-fusion modules to aggregate features and predict the segmentation masks. 
Our GL-Fusion method can reach 83.84\%, 81.76\% and 81.28\% performance in Dice score across three different views, with 10.49\%, 4.19\% and 4.68\% boosts when compared with the best single-view method~\cite{u-net}, and 4.75\%, 2.06\%, 3.57\% enhancement when compared to the best single-view with semi-supervised method CSS~\cite{weihang}. Also, compared with the different global fusion methods, our global and local fusion methods conduct significant improvements compared with the early-fusion approach. The visualization in Fig.~\ref{visualize} compares the segmentation quality with our GL-fusion method and others across three different views.
\subsection{Ablation Study}
In this section, we analyze the contribution to the performance of the proposed modules Multi-view Global Fusion Module (MGFM) and Multi-view Global Fusion Module (MGFM) of our framework. 
All results are illustrated in Table \textbf{2}. a-b, the baseline without adapting any fusion strategy presents the lowest average dice, while using only MGFM or MLFM module can boost the result to 80.20\%  and 78.41\%, respectively. 
The combination of these two modules can reach 82.29\% dice score with a 2.09\% increase in Dice score. In contrast, using the fusion method and cycle loss will lead to worse performance, while our proposed dense cycle loss can boost the result from 80.36\%  to 82.29\%.

\section{Conclusion}
In this paper, we propose a novel fusion framework called GL-Fusion, which jointly uses global and local information to enhance the segmentation performance of echocardiogram videos. Additionally, to ensure fair evaluation of the multi-view segmentation results, we introduce a multi-view echocardiogram video dataset called \textbf{MvEVD}, which provides full annotation for validating and testing performance. Our results demonstrate that the proposed GL-Fusion framework significantly outperforms other methods. In the future, we aim to further improve our method and make it more efficient.

\section*{Acknowledgements}
This work was partially supported by the Beijing Institute of Collaborative Innovation (BICI) under Grant HCIC-004, in collaboration with HKUST; the Foshan HKUST Projects under Grants FSUST21-HKUST10E and FSUST21-HKUST11E; and the Hong Kong Innovation and Technology Fund under Project ITS/030/21.

\bibliographystyle{splncs04}
\bibliography{cite}
\end{document}